\documentclass[10pt]{article}
\usepackage{abstract}
\usepackage{amsmath, amssymb}
\usepackage{graphicx}
\usepackage{caption}
\usepackage{authblk}
\usepackage{cite}
\usepackage{geometry}
\usepackage{multicol}
\newtheorem{definition}{Definition}
\usepackage{appendix}
\usepackage{changepage}
\usepackage{multirow}
\usepackage{makecell}
\usepackage{tcolorbox}

\title{\textbf{GraphXAIN: Narratives to Explain \\Graph Neural Networks}}
\author[1]{Mateusz Cedro\thanks{Corresponding author: mateusz.cedro@uantwerpen.be}}
\author[1]{David Martens}
\affil{University of Antwerp, Belgium}
\date{}

\begin{document}
\maketitle
\begin{multicols}{2}

\begin{adjustwidth}{-0.7cm}{0cm}
\noindent
\begin{minipage}[t]{0.57\textwidth}
\begin{abstract}
Graph Neural Networks (GNNs) are a powerful technique for machine learning on graph-structured data, yet they pose challenges in interpretability. Existing GNN explanation methods usually yield technical outputs, such as subgraphs and feature importance scores, that are difficult for non-data scientists to understand and thereby violate the purpose of explanations. Motivated by recent Explainable AI (XAI) research, we propose \textit{GraphXAIN}, a method that generates natural language narratives explaining GNN predictions. GraphXAIN is a model- and explainer-agnostic method that uses Large Language Models (LLMs) to translate explanatory subgraphs and feature importance scores into coherent, story-like explanations of GNN decision-making processes. Evaluations on real-world datasets demonstrate GraphXAIN's ability to improve graph explanations. A survey of machine learning researchers and practitioners reveals that GraphXAIN enhances four explainability dimensions: understandability, satisfaction, convincingness, and suitability for communicating model predictions. When combined with another graph explainer method, GraphXAIN further improves trustworthiness, insightfulness, confidence, and usability. Notably, 95\% of participants found GraphXAIN to be a valuable addition to the GNN explanation method. By incorporating natural language narratives, our approach serves both graph practitioners and non-expert users by providing clearer and more effective explanations.
\end{abstract}
\end{minipage}
\end{adjustwidth}    

\section{Introduction} 
The exponential growth in the complexity of machine learning models has led to architectures reaching billions of parameters, resulting in significant improvements in their performance \cite{LundbergL17, ribeiro2016should, guidotti_survey_2019, cedro2024blackboxcomplexdeep}. As these complex `black-box' models, characterized by their high accuracy yet lack of interpretability, continue to evolve, the demand for transparency has intensified \cite{molnar2022}. Explainable Artificial Intelligence (XAI) has emerged to address this challenge by enhancing the trustworthiness and transparency of complex decision-making processes \cite{Holzinger2022}.

\vspace{2em}
\begin{center}
    \includegraphics[width=1\columnwidth]{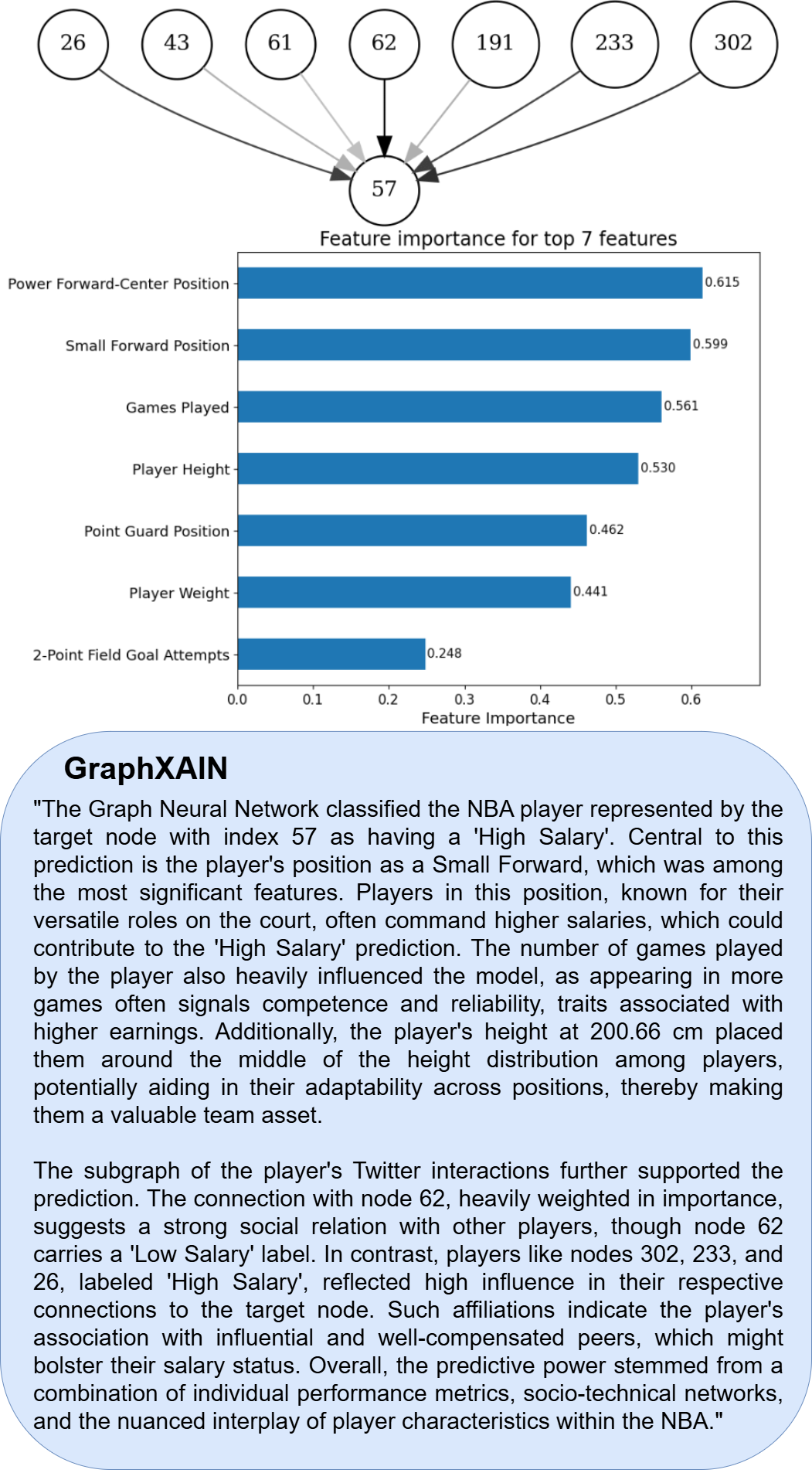}
    \captionof{figure}{LLM-generated GraphXAIN (bottom) complements GNNExplainer's\cite{gnnexplainergeneratingexplanationsgraph} subgraph and feature importance outputs (top) in explaining the GNN's prediction for the player 57 (node 57) in the NBA dataset, based on the player's attributes, field statistics, and social connections with other players (nodes).}
    \label{fig:XAIN_57}
\end{center}

\newpage
\begin{minipage}{\textwidth}
    \centering
    \includegraphics[width=1\linewidth]{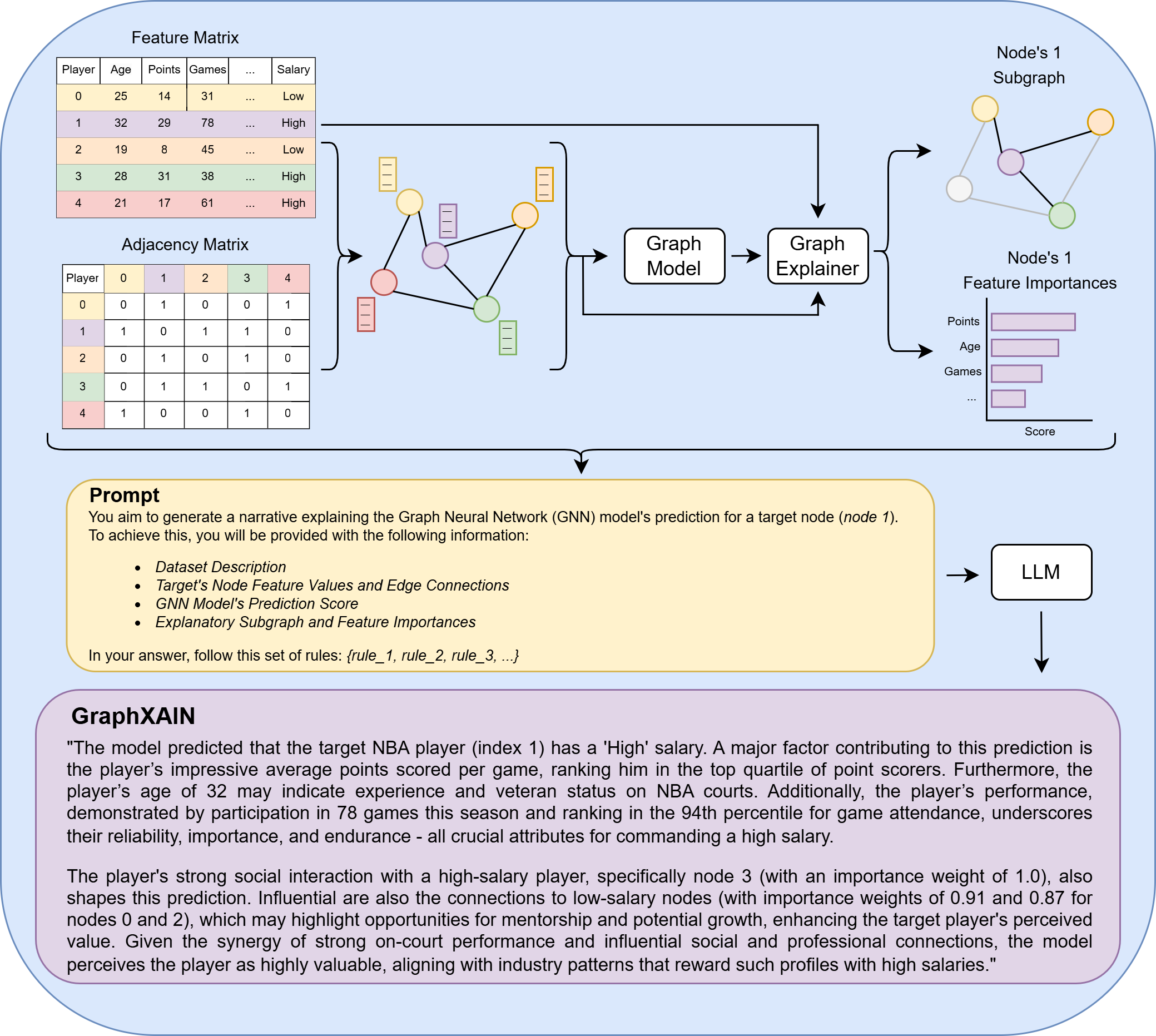}
    \captionsetup{justification=raggedright, singlelinecheck=false}
    \captionof{figure}{Workflow diagram of the GraphXAIN method. Graph-structured data, along with its corresponding features, is first processed by a GNN model. Next, a graph explainer generates an explanatory subgraph and corresponding feature importance values for a target node. The dataset description, target node features and edge connections, the GNN prediction score, the explanatory subgraph, and the feature importance values are then incorporated into a prompt. This prompt is processed by the LLM, which generates GraphXAIN, a complementary narrative to explain GNN's prediction.}
    \label{fig:workflow}
\end{minipage}
\vspace{1em}

Graph Neural Networks (GNNs) have recently gained notable success and become state-of-the-art solutions for modelling relational data characterized by nodes connected via edges \cite{Scarselli_2008, gat_veličković2018graphattentionnetworks, gin_xu2019powerfulgraphneuralnetworks, gcn_kipf2016semi}. However, the need for interpretability in GNNs remains \cite{gnnexplainergeneratingexplanationsgraph, cfgnnexplainercounterfactualexplanationsgraph, longa2024explainingexplainersgraphneural}.

Existing GNN explanation techniques predominantly offer explanatory subgraphs and feature-importance attributions, as shown in the upper section of Figure~\ref{fig:XAIN_57}. The figure illustrates an explanation produced by GNNExplainer for the GNN model’s classification of node 57 in the NBA dataset, taking into account the player’s attributes, field statistics, and social connections \cite{gnnexplainergeneratingexplanationsgraph}. However, the correct interpretation of the subgraph visualization alongside feature importance metrics alone can pose significant challenges for readers. Without a complementary narrative, practitioners employing this explanation technique to interpret the
\vfill
\vspace*{51.75em}
\noindent GNN's predictions must rely solely on subgraphs and feature importance outputs, which can be particularly challenging given the complex nature of GNN models. Despite its limitations, GNNExplainer remains a state-of-the-art graph explainer. Such an approach is not aligned with the \textit{comprehensibility postulate} introduced by Michalski\ (1983), which advocates that computer-generated results should be directly interpretable in natural language \cite{michalski_1983}. Moreover, effective explanations should enhance the alignment between the user's mental model and AI model \cite{martens2014explaining, kayande}. GNNExplainer's approach, among other graph explainers, does not adhere to these principles and exemplifies the phenomenon termed the ``\textit{Inmates Running the Asylum}", where solutions are technical and cater primarily to experts, overlooking the needs of less-technical practitioners who prefer natural-language explanations \cite{miller2017explainable, dahlstrom}.

Martens et al.\ (2023) proposed XAIstories, a framework that employs Large Language Models (LLMs) to generate narratives explaining the AI model's predictions based on SHAP and counterfactual (CF) explanations \cite{martens2024tellstorynarrativedrivenxai}. In their survey among data scientists and lay users, over 90\% of the general audience found these narratives convincing, while 83\% of the surveyed data scientists indicated they would likely use XAIstories to communicate explanations to non-expert audiences. By incorporating narrative communication alongside technical XAI methods, the model's decision-making process can be presented in a manner that aligns with human cognitive preferences, elevating explanations beyond mere descriptions and clarifying cause-and-effect relationships \cite{dahlstrom}.

To the best of our knowledge, no previous research proposes the general framework for natural language XAI Narratives to explain Graph Neural Network models' predictions. In this article, we introduce the first method to automatically generate \textit{GraphXAIN}, a natural language narrative, to explain the GNN model. By complementing explanatory subgraphs and feature importances with coherent XAI Narratives (see Figure \ref{fig:workflow}), we aim to further explain the decision-making process of GNNs in a more transparent and accessible way. We posit that this method will not only enhance interpretability but also facilitate more effective communication of model predictions across various graph applications.

To summarise, our main contributions are as follows:
\begin{itemize}
    \item We present a novel model-agnostic and explainer-agnostic method that generates natural language XAI Narratives to enhance the explainability of prediction models over graph-structured data.
    \item We illustrate our approach by integrating GraphXAIN with the existing graph XAI framework (GNNExplainer) for GNNs and demonstrating its explanatory abilities on real-world datasets, both in classification and regression tasks.
    \item We formalise the concepts of narrative and descriptive explanations within the context of XAI, clarifying their distinctions and discussing their implications for model interpretability.
    \end{itemize}

\section{Related Work}
\subsection{Explainability in Machine Learning}
Several approaches have been proposed to enhance the explainability of machine learning models across various modalities, including image data\cite{ribeiro2016should, simonyan_deep_2014, sundararajan_axiomatic_2017, adebayo_sanity_2020, hinns2024exposingimageclassifiershortcuts}, tabular data\cite{ribeiro2016should, LundbergL17, brughmans2022nicealgorithmnearestinstance}, natural language\cite{sundararajan_axiomatic_2017, LundbergL17, ribeiro2016should}, and unstructured data such as graphs\cite{gnnexplainergeneratingexplanationsgraph, cfgnnexplainercounterfactualexplanationsgraph}. Among the most popular XAI methodologies are post-hoc explanations, which aim to interpret prediction models after the training stage \cite{molnar2022}, while counterfactual explanations indicate what minimal changes in the input data are required to obtain a different predicted class \cite{martens2014explaining, wachter2018counterfactualexplanationsopeningblack}. These methods encompass feature importance measures, visualisation techniques, and surrogate models. For instance, SHAP (SHapley Additive exPlanations\cite{LundbergL17}) and LIME (Local Interpretable Model-agnostic Explanations\cite{ribeiro2016should}) estimate the contribution of each feature to a particular prediction.

However, despite these advancements, challenges persist in ensuring that explanations are both methodologically accurate and meaningful for a variety of stakeholders, not only for the data scientists \cite{goethals2023manipulationrisksexplainableai, martens2024tellstorynarrativedrivenxai, TAGex}. Explanations must bridge the gap between technical complexity and user comprehension, necessitating a careful balance between fidelity and interpretability\cite{dahlstrom, miller2017explainable}. The need for explainability methods that are understandable to less-technical users is primarily crucial in sensitive domains such as healthcare, finance, and legal systems, where understanding the model's decision-making process is essential for trust and transparency\cite{ribeiro2016should, molnar2022}.

\subsection{Explainability in Graph Neural Networks}
GNNs are increasingly used for modelling relational data in domains such as social networks, molecular structures, and knowledge graphs \cite{Scarselli_2008, gcn_kipf2016semi, gat_veličković2018graphattentionnetworks}. Their complex architectures, however, pose challenges for understanding and interpreting their predictions. The first method developed to address this issue is GNNExplainer \cite{gnnexplainergeneratingexplanationsgraph}, which provides instance-level explanations by identifying an explanatory subgraph and relevant node features which are most influential for a specific prediction. GNNExplainer formulates the explanation task as an optimisation problem, maximising the mutual information between the explanatory subgraph with a subset of node features and the original graph that is subject to explain. 

Among other graph explanation methods, Lucic et al.\ (2021) introduced CF-GNNExplainer\cite{cfgnnexplainercounterfactualexplanationsgraph}, which alters the GNNExplainer to answer 'what-if' questions using the counterfactual explanation approach. Rather than merely identifying influential features or subgraphs, CF-GNNExplainer searches for minimal perturbations to the original graph that would alter the GNN model's prediction by edge deletion. This method demonstrates how small changes in graph structure impact outcomes, enhancing understanding of the model's decision-making process.

Although the aforementioned state-of-the-art graph explanation frameworks are methodologically sound, they merely provide users with additional graphs and node feature importance values that are not easily interpretable (see Figure \ref{fig:XAIN_57}), thereby limiting their practical utility and compelling practitioners to construct the explanatory narrative themselves.

We posit that incorporating natural language into GNN explanations could bridge the gap between technical outputs and human understanding by translating complex model reasoning into accessible narratives, thereby enhancing comprehension and trust among practitioners. However, previous methods to generate them are not tailored to the popular GNNExplainer output and/or provide descriptive explanation rather than a cohesive narrative (see below for a discussion of the important differences). Giorgi et al.\ (2024) addressed this issue by using LLMs to generate textual explanations for counterfactual graphs\cite{giorgi2024naturallanguagecounterfactualexplanations}. However, their explanations lack contextual information and do not illustrate cause-and-effect relationships, resulting in primarily descriptive communication rather than narrative explanation - the latter being more valuable for conveying more complex information\cite{dahlstrom} (see Appendix~\ref{app:Giorgi} for examples).

He et al.\ (2024) used LLMs to generate natural language explanations for counterfactual graphs in the context of molecular property prediction\cite{he2024explaininggraphneuralnetworks}. However, since their framework adheres to domain-specific knowledge, it cannot be considered a general method for graph model explanations with natural language explanations. Furthermore, the explanations produced by their method exhibit the same limitation as in Giorgi et al.\ (2024), being more descriptive rather than narrative (see Appendix~\ref{app:He} for examples).

The most comparable approach to ours is presented by Pan et al.\ (2024), who developed TAGExplainer, a method for generating natural language explanations for Text-Attributed Graphs (TAGs)\cite{TAGex}. Although their explanations incorporate elements of narrative communication, their framework is limited only to TAGs. This limitation restricts the generalisability of TAGExplainer to broader graph data modelling contexts, as TAGs represent only a subset of real-world graph data (see Appendix~\ref{app:Pan} for examples).

\section{Methods}
\subsection{XAI Narrative and Description}
Research in psychology and communication theory indicates that narrative-based explanations are more accessible and memorable than descriptive methods, making them effective for conveying scientific evidence to non-expert audiences \cite{dahlstrom, dahlstrom2018escaping, biecekposition}. Moreover, narratives are processed more rapidly by individuals without prior knowledge and are more engaging and persuasive, thereby enhancing trust and understanding of AI models \cite{miller2017explainable, martens2024tellstorynarrativedrivenxai, biecekposition}.

Further, as narrative communication relies on contextual cause-and-effect relationships, it is considerably more challenging to fragment a narrative into smaller, meaningful segments without either significantly altering the interpretation of these segments or disrupting the coherence of the original narrative \cite{dahlstrom}. Consequently, narratives are often perceived as storytelling, characterised by a coherent structure comprising an introduction, main body, and conclusion. In contrast, descriptive, context-free communication can be readily fragmented into smaller units while still effectively conveying the necessary information \cite{dahlstrom}. The XAIstories method addresses aforementioned explanation limitations by enhancing the narrative communication of SHAP and CF explanations, aligning with the research on human-AI interactions \cite{martens2024tellstorynarrativedrivenxai, miller2017explainable, biecekposition}.

Having identified the need to distinguish between narrative and descriptive explanations in XAI, and drawing on social science research \cite{grice1975logic, dahlstrom, miller2017explainable, dahlstrom2018escaping}, we propose definitions for both terms:

\begin{definition}[XAI Narrative]
A XAI Narrative provides a structured, story-like representation of a model's prediction. Narrative explanations illustrate the relationships between key features in a context-dependent manner, providing a coherent and comprehensive understanding of how the model arrives at specific outcomes.
\end{definition}

\begin{definition}[XAI Description]
A XAI Description provides a static presentation of key features or attributes relevant to a model's prediction, delivered in a context-free and fact-based manner.
\end{definition}

In the context of explaining GNN models, a description would just list the most important features and neighbouring nodes. This relates closely to the data-to-text or graph-to-text approaches\cite{puduppully2019, gatt2018}. Figure \ref{fig:XAI_57_description} presents an example of XAI Description for the subgraph and feature importance output provided by GNNExplainer shown in Figure \ref{fig:XAIN_57}. Clearly, in our context, a XAI Description is less valuable than an XAI Narrative, as descriptions are less accessible and memorable than narrative communication methods, making them less effective for conveying scientific evidence to a broader audience \cite{dahlstrom, martens2024tellstorynarrativedrivenxai}.
 
\subsection{From GNNs to Natural Language Narratives}
To address the issue of purely technical explanations for GNN predictions, we propose GraphXAINs, natural language narratives for graphs. We propose the following definition of GraphXAIN, a XAI Narratives for graphs:

\begin{definition}[GraphXAIN] A GraphXAIN is a XAI Narrative tailored for graph-structured data.
\end{definition}

Our solution involves converting subgraph structures and feature importance metrics derived from graph explainers into GraphXAINs - coherent natural language narratives. The workflow, presented in Figure \ref{fig:workflow}, begins with the training of a GNN model on a relational graph dataset for classification or regression tasks. Our approach is not limited to pure graphs; data can include associated node and edge features. Further, using the graph explainer method, we extract relevant subgraphs and feature importances. The subgraph, along with feature importance, are then transformed and prompted into an LLM for further inference. The LLM is guided to generate a natural language narrative that explains the prediction of the target node's label or score, highlighting the contributions of specific neighbouring nodes, corresponding edges, and node features in a context-dependent manner, resulting in a coherent cause-and-effect narrative presented to the end user. 

Importantly, our framework is agnostic to the graph data type, graph model, performed task (classification and regression), and graph explainer, allowing its application across various graph scenarios and applications. By translating technical explanations into coherent narratives, we aim to make graph model explanations more intuitive and accessible for practitioners.

\section{Experiments}
\subsection{Datasets}
We conduct our experiments on two real-world graph datasets, one used for the node classification task and the other for the node regression task. The dataset for the node classification task, NBA players' dataset\footnote{https://www.kaggle.com/noahgift/social-power-nba}, includes players' performance statistics from one NBA season, alongside various personal attributes such as height, weight, age, and nationality. The graph is constructed by linking NBA players based on their relationships and interactions on the Twitter platform. After preprocessing, filtering out disconnected nodes, and ensuring the graph is undirected, the final graph comprised of the total number of 400 nodes (players), 42 node features, and 21,242 edges (Twitter connections). The node classification task involves predicting whether a player's salary exceeds the median salary.

The dataset used to perform node regression task, IMDB movie dataset\footnote{https://www.kaggle.com/datasets/harshitshankhdhar/imdb-dataset-of-top-1000-movies-and-tv-shows}, includes information on 1,000 movies from the IMDB database. The dataset includes information such as movie director, main actors, release year, duration, and genre. The graph is created by connecting nodes, represented by a single movie, by edges which link two movies if at least one actor played in both movies. In total, the graph consists of 1,000 nodes and 5,608 undirected edges. Each node has 12 features. The node regression task involves predicting the IMDB rating score ranging between 0 and 10, where 10 is the maximum score indicating the general appreciation of the movie by the audience.

\subsection{Graph Models}
Prior to training, both datasets were randomly divided into separate training, validation, and test sets in a 60/20/20 split. The GNN classification model consists of two Graph Convolutional Network (GCN) layers \cite{gcn_kipf2016semi}, with 16 hidden channels. For training, we use the Binary Cross-Entropy loss function with the AdamW \cite{loshchilov2019decoupledweightdecayregularization} optimizer with a learning rate of 0.001 and a weight decay of 5$\times$10$^{-4}$. The training process continued for 1,400 epochs, resulting in the GCN classification model that achieved a test AUC of 0.80.

The GNN regressor model consists of two GCN layers with 32 hidden channels. The Root Mean Square Error loss function is used to train the model with the AdamW optimizer with a learning rate of 0.01 and a weight decay of 5$\times$10$^{-4}$. The training process continues for 7,500 epochs, with the early stopping of 500 steps, resulting in the RMSE of 0.28 on the held-out test set.

\begin{center}
    \includegraphics[width=1\columnwidth]{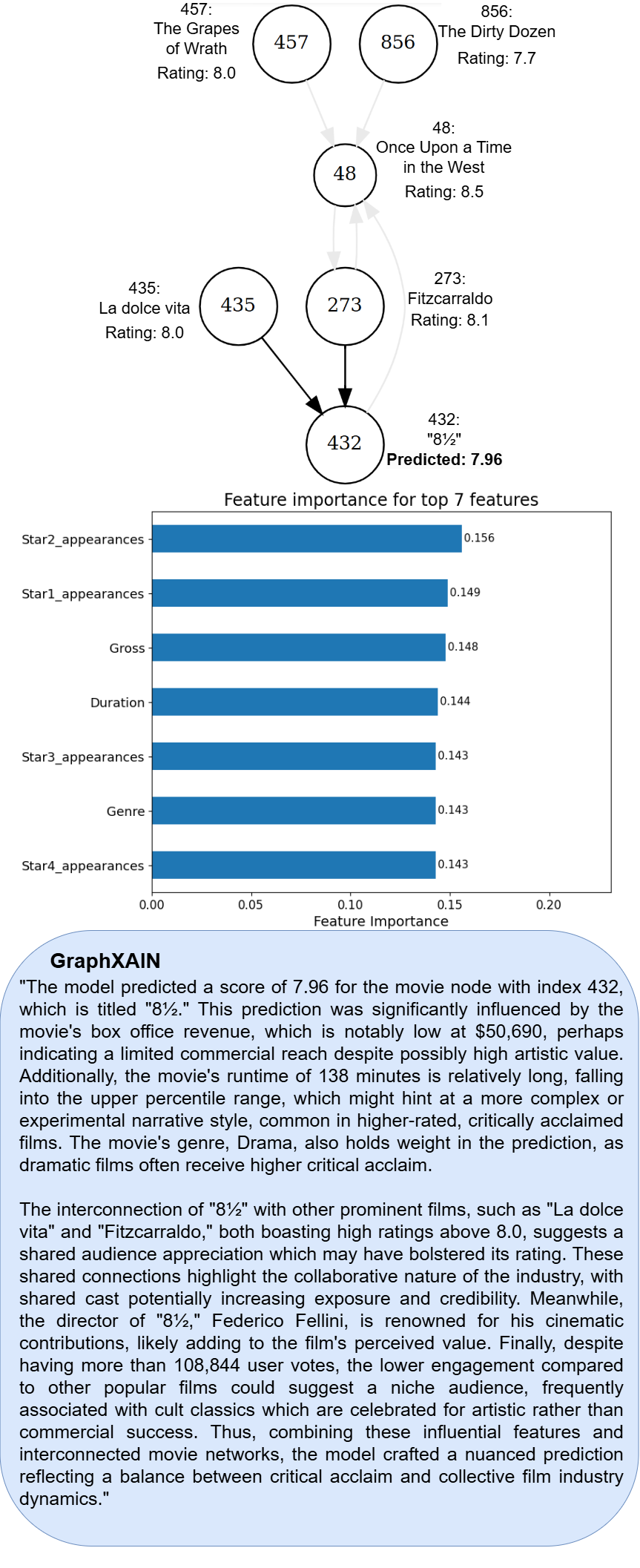}
    \captionof{figure}{LLM-generated GraphXAIN (bottom) complements GNNExplainer's \cite{gnnexplainergeneratingexplanationsgraph} subgraph and feature importance outputs (top) in explaining the GNN’s prediction for the movie ``$8\frac{1}{2}$" (node 432) in the IMDB dataset, based on movie features and connections with other movies (nodes) via shared actors.}
    \label{fig:XAIN_432_reg}
\end{center}
\subsection{Graph Explainer}
To obtain subgraph and feature importance explanations, we use GNNExplainer \cite{gnnexplainergeneratingexplanationsgraph}, the current state-of-the-art method for explaining GNN model predictions. The GNNExplainer formulates the explanation task as an optimization problem that maximizes the mutual information between the explanatory subgraph and a subset of node features, relative to the input graph under consideration. In both scenarios (classification and regression), the GNNExplainer training process was conducted for 200 epochs, adhering to the default settings recommended for this explainer by the authors\cite{gnnexplainergeneratingexplanationsgraph}. Different shades of the edges in the subgraph indicate the strength of importance of each connection, with darker edges indicating higher strength. Nevertheless, in our framework, any graph explainer may be used as long as it provides an explanatory subgraph and feature importance values.

\subsection{Graphs and Large Language Models}
XAI Narratives are derived from the GNNExplainer output, with the use of a LLM (GPT-4o). Using a specifically developed prompt, we guide the LLM in generating the GraphXAIN that accurately reflects the graph model's prediction. The language model is provided with a prompt containing the target node's feature values, along with feature importance and the subgraph derived from GNNExplainer. 

Additionally, feature values of the nodes within the subgraph and the final GNN model prediction for the target node are included. In our visualisations, we use the seven most important features and restrict the subgraph to seven nodes, aligning with Miller's\ (1956) \cite{miller1956} theory on cognitive limits, which suggests that seven pieces of information represent an optimal amount for receiving, processing and recalling. The full prompt schema used to generate GraphXAINs is presented in Appendix~\ref{app:prompt}.

\section{Results}
In the following, we provide examples presenting various automatically generated GraphXAINs to explain the GNN model’s prediction for the target node. It is important to clarify that the presented results are not selectively chosen - rather, a subset of nodes (subject to explanation) is randomly drawn from both datasets. The GraphXAINs are subsequently generated for each of the selected nodes. Figure \ref{fig:XAIN_57} and Figure \ref{fig:XAIN_432_reg} present examples of generated GraphXAINs for node classification and node regression tasks, respectively. For some other GraphXAIN examples, see Appendix~\ref{app:Cedro}.

The results of the GraphXAINs show the effectiveness of generating XAI narrative-based explanations from the context-free subgraphs and feature importance values. Unlike other GNN XAI descriptive outputs, which typically present static feature values, the outputs generated by our method provide a coherent XAI Narrative. Not only do these narratives articulate which features contributed most to the model's prediction, but also explain how and why these features combined in a cause-and-effect manner. For instance, the narrative for a high-salary prediction highlights the player's field and performance statistics, such as position and the number of games played, while crucially contextualising these within broader patterns of team dynamics and social interactions. 

Figure \ref{fig:XAI_57_description} presents a possible XAI Description for the GNN model’s final prediction of node 57, outlining key features contributing to the target label prediction. The XAI Description is presented in a static, context-free, and non-coherent format, emphasising the fundamental distinctions between a description and a narrative. The advantage of the XAI Narrative presented at bottom of the Figure \ref{fig:XAIN_57} over the provided XAI Description is evident. Thus GraphXAIN goes beyond the XAI Description as shown in Figure \ref{fig:XAI_57_description}.

\begin{center}
    \includegraphics[width=1\columnwidth]{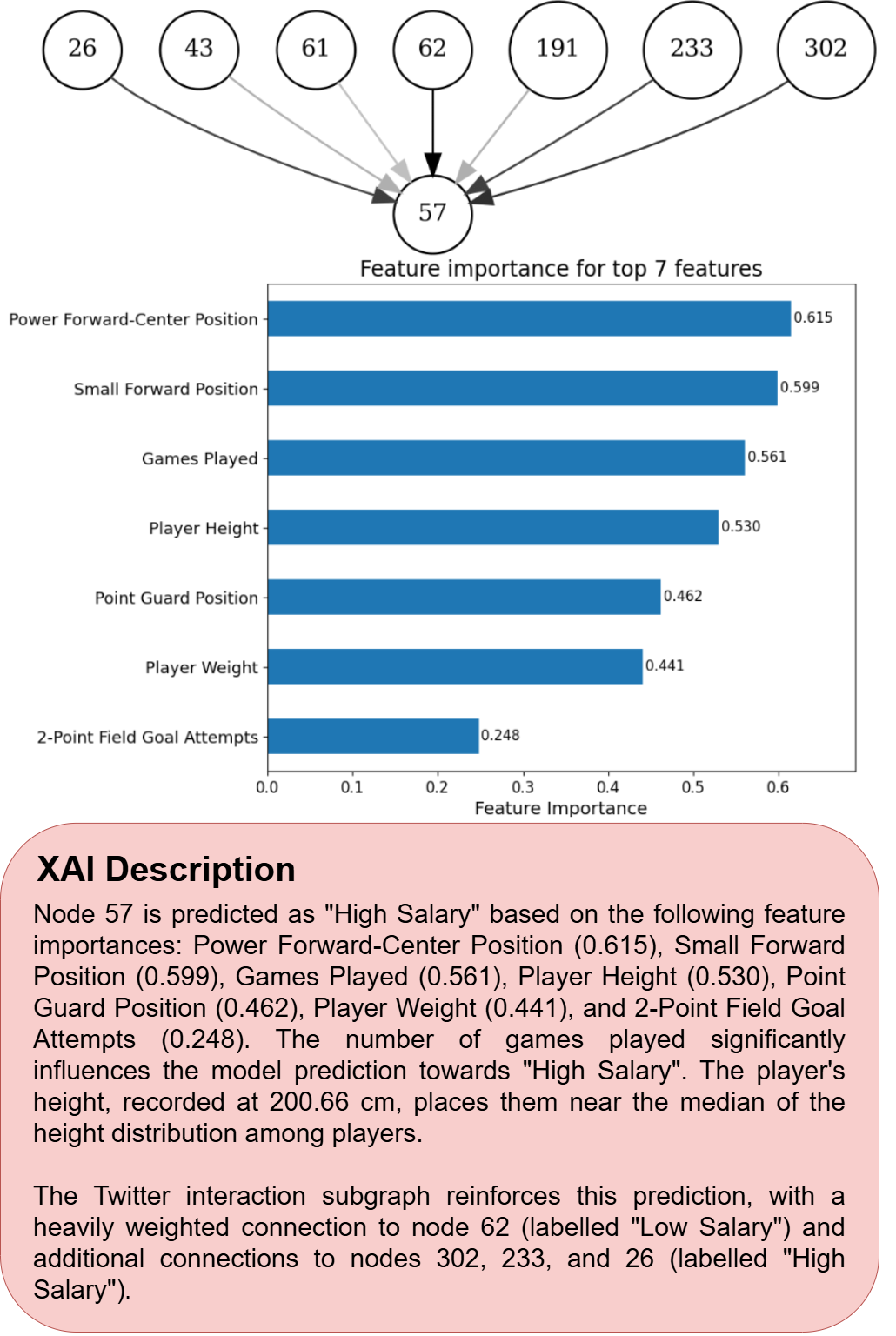}
    \captionof{figure}{LLM-generated XAI Description (bottom) for the GNNExplainer's \cite{gnnexplainergeneratingexplanationsgraph} subgraph and feature importance outputs (top) to explain the GNN's prediction for player (node) 57. Compared to GraphXAIN's output, XAI Description focuses on features and connections in a static and context-free format, which is less valuable than a XAI Narrative.}
    \label{fig:XAI_57_description}
\end{center}

\end{multicols}
\begin{table}[t]
\begin{minipage}{\textwidth}
\centering
\begin{tabular}{l l}
\hline
Dimension & Question \\ \hline
Understandability & Q1: I find the explanation understandable. \\
Trustworthiness & Q2: This explanation increased my trust in the model’s prediction. \\
Insightfulness & \makecell[l]{Q3: The explanation helped me gain insight into the factors influencing \\  \hspace{0.725cm}the classification/regression.} \\ 
Satisfaction & Q4: I am satisfied with the explanation’s clarity and thoroughness. \\
Confidence & Q5: I feel confident in the classification/regression after reviewing the explanation. \\
Convincingness & \makecell[l]{Q6: I find this explanation convincing in justifying \\ \hspace{0.725cm}why node \textit{X} is classified as Y/predicted to have a regression score of Y.} \\
Communicability & \makecell[l]{Q7: I find this explanation method suitable for communicating \\ \hspace{0.725cm}the model’s predictions to others.} \\
Usability & \makecell[l]{Q8: If I work with Graph Neural Network models in the future, \\ \hspace{0.725cm}I am likely to use this explanation method to explain the model's prediction.} \\
\hline
\end{tabular}
\captionof{table}{Survey Questions for Assessing Explainable AI Method Perceptions}
\label{tab:questions}
\end{minipage}
\end{table}

\begin{multicols}{2}
This approach ensures that the explanation is not limited to mere numerical and visual descriptions, but instead offers a comprehensive, story-like narrative that enables the practitioners to understand the model's decision-making process more intuitively. By conveying the reasoning behind the prediction, the XAI Narrative explanation addresses a primary goal in XAI by bridging the gap between technical model outputs and user comprehension, thereby reflecting both the principles of narrative communication and the general objectives of explainability research \cite{dahlstrom}.

Note that in the IMDB dataset, movie titles are available and directly mapped to nodes, whereas in the NBA dataset, player names are hashed, preventing their direct mapping. We hypothesize that having actual names would allow GraphXAIN to produce more effective explanations by drawing on specific player information.

\section{Evaluation with human \\ subjects: a user study}
To evaluate the proposed GraphXAIN method, we conducted a user study similar to Baniecki et al.\ (2023) and Martens et al.\ (2024)\cite{baniecki2024grammar, martens2024tellstorynarrativedrivenxai}. Twenty active researchers and practitioners from the field of machine learning (ML) and data science (DS) participated in the evaluation survey. Seventy-five percent of the participants are academic ML researchers, remaining 25\% are industry ML/DS practitioners. Regarding professional experience, 30\% of the participants had between 0 and 2 years of experience, 50\% had 3 to 5 years, 15\% had 6 to 10 years, and 5\% had more than 10 years of experience in the domain.

The purpose of the survey was to evaluate the impact of the proposed GraphXAIN method in comparison with the current state-of-the-art XAI method\cite{gnnexplainergeneratingexplanationsgraph}
to explain GNN predictions. Six randomized GNN explanation examples were shown to each participant sequentially. Immediately after viewing each XAI method, participants are asked to rate their agreement with eight questions on a five-point Likert scale, ranging from “Strongly disagree” (1) to “Strongly agree” (5), where higher scores indicate stronger agreement or preference. Each participant was presented with three examples drawn from both classification and regression tasks: 1) a current state-of-the-art GNNExplainer \cite{gnnexplainergeneratingexplanationsgraph} explanation comprising a subgraph and feature-importance attributions, 2) GraphXAIN, which provides a narrative explanation of the model’s prediction, 3) a combined approach that integrates GNNExplainer and GraphXAIN. The order of presenting the three aforementioned methods to the participants was randomized.

Respondents evaluated XAI methods across eight dimensions: understandability, trustworthiness, insightfulness, satisfaction, confidence, convincingness, communicability, and usability. Table \ref{tab:questions} presents the survey questions used to measure each dimension. The following section presents the theoretical principles for selecting these evaluation dimensions.

\begin{itemize}
    \item Understandability -  In Hoffman et al.\ (2018), Miller\ (2019), and Mohseni et al.\ (2021), the understanding of XAI explanation is declared to be a crucial part of the XAI methods to ensure understandability of the rationale behind the model’s predictions \cite{miller2019explanation, mohseni2021multidisciplinary, hoffman2018metrics}.
    \item Trustworthiness - Hoffman et al.\ (2018) note that fostering appropriate trust is one of the core objectives of explanations, underscoring the need to assess whether an explanation effectively increases trust \cite{hoffman2018metrics}.
    \item Insightfulness - Riberio et al.\ (2016) declare that a good explanation should provide additional insights into model predictions \cite{ribeiro2016should}.
    \item Satisfaction - Miller\ (2019) investigates the cognitive and social dimensions that render explanations meaningful to humans, positing that, among others, practitioners and user satisfaction is fundamental to effective explanatory strategies \cite{miller2019explanation}.
    \item Confidence - Ribeiro et al.\ (2024) argue that explanations should enable individuals to gain confidence in the model \cite{ribeiro2024reliable}.
    \item Convincingness - Miller (2019) underscores that an explanation must be persuasive to the user \cite{miller2019explanation}.
\end{itemize}

\end{multicols}
\begin{table}[t]
\centering
\begin{minipage}{\textwidth}
\begin{tabular}{l l l l l l}
\hline
\multirow{2}{*}{Dimension} &
\multirow{2}{*}{Method Comparison ($M_1$ vs $M_2$)} &
\multirow{2}{*}{$M_1$$\uparrow$} &
\multirow{2}{*}{$M_2$$\uparrow$} &
\multirow{2}{*}{$\Delta (M_2,M_1$)} &
\multirow{2}{*}{$p$-value} \\
& & & & & \\
\hline

\multirow{3}{*}{Understandability}
 & SubG+FI vs \textbf{GraphXAIN}            
   & $3.3_{\scriptstyle \pm 0.8}$ & $4.3_{\scriptstyle \pm 0.5}$ & $0.95_{\scriptstyle \pm 1.02}$ & 0.004* \\
 & SubG+FI vs \textbf{SubG+FI+GraphXAIN} 
    & $3.3_{\scriptstyle \pm 0.8}$ & $4.4_{\scriptstyle \pm 0.6}$ & $1.05_{\scriptstyle \pm 0.94}$ & 0.003* \\
 & GraphXAIN vs SubG+FI+GraphXAIN    
   & $4.3_{\scriptstyle \pm 0.5}$ & $4.4_{\scriptstyle \pm 0.6}$ & $0.1_{\scriptstyle \pm 0.58}$ & 1.000 \\
\hline

\multirow{3}{*}{Trustworthiness}
 & SubG+FI vs GraphXAIN              
   & $3.3_{\scriptstyle \pm 0.7}$ & $3.8_{\scriptstyle \pm 0.6}$ & $0.42_{\scriptstyle \pm 1.03}$ & 0.254 \\
 & SubG+FI vs \textbf{SubG+FI+GraphXAIN} 
    & $3.3_{\scriptstyle \pm 0.7}$ & $4.2_{\scriptstyle \pm 0.6}$ & $0.92_{\scriptstyle \pm 0.94}$ & 0.005* \\
 & GraphXAIN vs SubG+FI+GraphXAIN    
   & $3.8_{\scriptstyle \pm 0.6}$ & $4.2_{\scriptstyle \pm 0.6}$ & $0.5_{\scriptstyle \pm 0.84}$ & 0.072 \\
\hline

\multirow{3}{*}{Insightfulness}
 & SubG+FI vs GraphXAIN              
   & $3.9_{\scriptstyle \pm 0.6}$ & $4.2_{\scriptstyle \pm 0.4}$ & $0.28_{\scriptstyle \pm 0.62}$ & 0.178 \\
 & SubG+FI vs \textbf{SubG+FI+GraphXAIN} 
    & $3.9_{\scriptstyle \pm 0.6}$ & $4.4_{\scriptstyle \pm 0.6}$ & $0.5_{\scriptstyle \pm 0.63}$ & 0.018* \\
 & GraphXAIN vs SubG+FI+GraphXAIN    
   & $4.2_{\scriptstyle \pm 0.4}$ & $4.4_{\scriptstyle \pm 0.6}$ & $0.22_{\scriptstyle \pm 0.50}$ & 0.251 \\
\hline

\multirow{3}{*}{Satisfaction}
 & SubG+FI vs \textbf{GraphXAIN}              
   & $3.0_{\scriptstyle \pm 1.1}$ & $3.8_{\scriptstyle \pm 0.6}$ & $0.85_{\scriptstyle \pm 1.29}$ & 0.045* \\
 & SubG+FI vs \textbf{SubG+FI+GraphXAIN} 
    & $3.0_{\scriptstyle \pm 1.1}$ & $4.0_{\scriptstyle \pm 0.6}$ & $1.05_{\scriptstyle \pm 1.0}$ & 0.003* \\
 & GraphXAIN vs SubG+FI+GraphXAIN    
   & $3.8_{\scriptstyle \pm 0.6}$ & $4.0_{\scriptstyle \pm 0.6}$ & $0.2_{\scriptstyle \pm 0.70}$ & 0.586 \\
\hline

\multirow{3}{*}{Confidence}
 & SubG+FI vs GraphXAIN              
   & $3.0_{\scriptstyle \pm 0.9}$ & $3.6_{\scriptstyle \pm 0.7}$ & $0.62_{\scriptstyle \pm 1.09}$ & 0.085 \\
 & SubG+FI vs \textbf{SubG+FI+GraphXAIN} 
    & $3.0_{\scriptstyle \pm 0.9}$ & $3.9_{\scriptstyle \pm 0.8}$ & $0.88_{\scriptstyle \pm 1.17}$ & 0.015* \\
 & GraphXAIN vs SubG+FI+GraphXAIN    
   & $3.6_{\scriptstyle \pm 0.7}$ & $3.9_{\scriptstyle \pm 0.8}$ & $0.25_{\scriptstyle \pm 1.02}$ & 0.786 \\
\hline

\multirow{3}{*}{Convincingness}
 & SubG+FI vs \textbf{GraphXAIN}              
   & $3.2_{\scriptstyle \pm 0.7}$ & $3.8_{\scriptstyle \pm 0.8}$ & $0.62_{\scriptstyle \pm 0.96}$ & 0.030* \\
 & SubG+FI vs \textbf{SubG+FI+GraphXAIN} 
    & $3.2_{\scriptstyle \pm 0.7}$ & $4.1_{\scriptstyle \pm 1.0}$ & $0.9_{\scriptstyle \pm 1.15}$ & 0.021* \\
 & GraphXAIN vs SubG+FI+GraphXAIN    
   & $3.8_{\scriptstyle \pm 0.8}$ & $4.1_{\scriptstyle \pm 1.0}$ & $0.28_{\scriptstyle \pm 1.18}$ & 0.314 \\
\hline

\multirow{3}{*}{Communicability}
 & SubG+FI vs \textbf{GraphXAIN}              
   & $2.8_{\scriptstyle \pm 1.0}$ & $3.8_{\scriptstyle \pm 0.8}$ & $1.0_{\scriptstyle \pm 1.39}$ & 0.020* \\
 & SubG+FI vs \textbf{SubG+FI+GraphXAIN} 
    & $2.8_{\scriptstyle \pm 1.0}$ & $4.2_{\scriptstyle \pm 0.7}$ & $1.45_{\scriptstyle \pm 1.22}$ & 0.001* \\
 & GraphXAIN vs \textbf{SubG+FI+GraphXAIN}    
   & $3.8_{\scriptstyle \pm 0.8}$ & $4.2_{\scriptstyle \pm 0.7}$ & $0.45_{\scriptstyle \pm 0.63}$ & 0.019* \\
\hline

\multirow{3}{*}{Usability}
 & SubG+FI vs GraphXAIN              
   & $3.2_{\scriptstyle \pm 0.9}$ & $3.6_{\scriptstyle \pm 1.0}$ & $0.35_{\scriptstyle \pm 1.43}$ & 0.284 \\
 & SubG+FI vs \textbf{SubG+FI+GraphXAIN} 
    & $3.2_{\scriptstyle \pm 0.9}$ & $4.0_{\scriptstyle \pm 0.8}$ & $0.78_{\scriptstyle \pm 0.9}$ & 0.008* \\
 & GraphXAIN vs SubG+FI+GraphXAIN    
   & $3.6_{\scriptstyle \pm 1.0}$ & $4.0_{\scriptstyle \pm 0.8}$ & $0.42_{\scriptstyle \pm 1.24}$ & 0.541 \\
\hline

\end{tabular}
\captionof{table}{Pairwise post-hoc Wilcoxon signed-rank tests comparing XAI methods for each dimension question. Columns $M_1$ and $M_2$ report the mean ± standard deviation (SD) of participants’ preferences for each method. Higher values indicate increased preference; responses were recorded on a 5-point Likert scale with the following options: Strongly disagree, Somewhat disagree, Neither agree nor disagree, Somewhat agree, and Strongly agree (corresponding to 1 through 5, respectively, with 5 indicating the highest preference). The value \(\Delta(M_2, M_1)\) represents the mean difference (with SD) between the corresponding methods. Bonferroni-corrected p-values (actual p-values multiplied by the total number of comparisons) for differences between the assessed methods are provided. Note:* indicates statistically significant differences at \(\alpha = 0.05\). Bolded methods indicate statistically significant preferences.}
\label{tab:posthoc_results}
\end{minipage}
\end{table}

\begin{multicols}{2}
\begin{itemize}
    \item Communicability - Miller (2019) argues that the information exchanged between the explainer and the explainee should align with the general rules of cooperative conversation, proposed by Grice’s\ (1975) \cite{grice1975logic}, ensuring that it remains relevant to the explainee’s context and builds upon their prior knowledge \cite{miller2019explanation}.
    \item Usability - Miller (2017) argues that individuals evaluate explanations according to pragmatic influences of causality, which encompass criteria such as usefulness \cite{miller2017explainable}.
\end{itemize}

Table \ref{tab:posthoc_results} presents the survey results along with the statistical significance from Wilcoxon signed-rank tests assessing the differences between XAI methods, conducted at the  \(\alpha = 0.05\) significance level.

When comparing preferences between GNNExplainer (presented in Table \ref{tab:posthoc_results} as “SubG+FI”) and GraphXAIN, the latter is preferred in four dimensions: understandability, satisfaction, convincingness, and communicability, while no statistically significant differences are observed in the other four dimensions.

In another comparison, when evaluating the GraphXAIN alone against a combined explanation that integrates GNNExplainer's outputs with the GraphXAIN narrative (SubG+FI+GraphXAIN), only one dimension, communicability, shows a statistically significant difference, with the combined method being preferred. For the remaining seven dimensions, no significant differences occur.

Moreover, across all eight investigated dimensions, the combined method (SubG+FI+GraphXAIN) is preferred over the GNNExplainer method (SubG+FI). This finding suggests that incorporating GraphXAIN’s narrative component into the technical subgraph and feature importance explanations consistently enhances the overall quality of the explanation.

Furthermore, 95\% of participants (19/20) answered “Yes” to the question, “Do you think that the narratives are a useful addition to explaining the GNN model's predictions?”. This response aligns with the observed advantages of the combined method (SubG+FI+GraphXAIN) over the standalone GNNExplainer outputs.

Additionally, at the \(\alpha=0.05\) significance level, no statistically significant differences are identified between responses for the NBA and IMDb datasets across any condition or question. Overall, these findings indicate that participants’ ratings do not differ meaningfully between the two datasets, which represent distinct graph predictive tasks.

Although the sample size is modest, the survey findings demonstrate that GraphXAIN consistently outperforms the current state-of-the-art graph explanation method, serving as a valuable enhancement to improve understanding, trust, insightfulness, satisfaction, confidence, convincingness, communicability, and usability.

\section{Conclusion and Future Work}
In this work, we introduced GraphXAIN, a novel solution that transforms explanatory subgraphs and feature importance metrics into coherent narrative explanations for Graph Neural Network (GNN) models. \newline GraphXAIN is a model-agnostic and explainer-agnostic approach that complements traditional graph explainer outputs by leveraging Large Language Models (LLMs). Results obtained from generating narrative explanations for both node classification and node regression tasks on real-world datasets demonstrate that these narratives provide clearer insights into the decision-making processes of complex GNN models, making them more accessible to non-expert users. In doing so, our method addresses a significant gap in the current Explainable AI landscape by moving beyond technical, descriptive outputs and offering intuitive, story-like explanations that enhance practitioners’ comprehension and trust.

A survey conducted among machine learning researchers and practitioners highlights the advantages of GraphXAIN over GNNExplainer \cite{gnnexplainergeneratingexplanationsgraph}. At the $\alpha=0.05$ level, the GraphXAIN method significantly improved understandability, satisfaction, convincingness, and suitability for communicating the model's predictions. Furthermore, when GNNExplainer’s outputs are combined with GraphXAIN narratives, all eight measured dimensions (understandability, trustworthiness, insightfulness, satisfaction, confidence, convincingness, communicability, and future usability) show significant improvement compared to using GNNExplainer outputs alone, indicating that integration of the XAI Narratives with subgraph and feature importance outputs enhances users’ perceptions of the explanation. Notably, 95\% of participants indicate that GraphXAIN is a valuable addition to the explanation of GNN predictions.

In future work, employing quantitative metrics will be essential for a more thorough assessment of the generated narrative explanations \cite{ichmoukhamedov2024good}. One promising direction is to tailor our method to generate graph counterfactual narratives that describe the minimal change required to alter the classifier's decision, similar to the approach proposed by Martens et al.\ (2024) \cite{martens2024tellstorynarrativedrivenxai}. Moreover, due to the unstructured nature of graph data, LLMs may be more prone to generating hallucinations since they struggle to interpret unstructured inputs as accurately as structured data \cite{hallucinationlargelanguage}. This challenge can be mitigated by integrating our approach with methods specifically designed for graph inputs, such as G-Retriever \cite{gretrieverretrievalaugmentedgenerationtextual}.

\section*{Data and Code Availability}
The data and code used in this research, as well as the survey results, are available at: \newline \texttt{https://github.com/ADMAntwerp/GraphXAIN}

\section*{Acknowledgments}
This research received funding from the Flemish Government under the “Onderzoeksprogramma Artifici\"{e}le Intelligentie (AI) Vlaanderen”.

\bibliographystyle{plain}
\bibliography{references}
\end{multicols}

\newpage
\appendix
\section{Prompt Schema}
\label{app:prompt}

Figure \ref{fig:prompt} and Figure \ref{fig:prompt_cd} present the prompt schema and the rules used to generate GraphXAINs.

\begin{figure}[htbp]
  \centering
\begin{tcolorbox}[colback=blue!5, colframe=blue!80, title=GraphXAIN Prompt, sharp corners]
Your goal is to generate a textual explanation or narrative explaining why a graph explainer produced a certain target node's explanation subgraph and feature importance for a Graph Neural Network (GNN) model's prediction of a target node instance.

To achieve this, you will be provided with the following information:

\begin{itemize}
    \item \textbf{Machine Learning Task Information:} Details about the machine learning task.
    \item \textbf{Dataset Information:} Information about the dataset used.
    \item \textbf{Target Node's Data:} Information about target node's continuous features: feature name, feature importance, feature continuous value, percentile rank, and feature description.
    \item \textbf{Target Node's Categorical Data:} Information about target node's categorical/binary features: feature name, feature importance, feature binary value, share of positive class in the dataset that has that feature present, and feature description.
    \item \textbf{Target Node's Number of Edges and Labels Ratio:} Dictionary with the total number of edges connected to the target node and the ratio of class labels among its neighbouring nodes.
    \item \textbf{Target Node's Feature Importance:} List of two-element tuples containing feature names and their importance scores of all of the target node's features sorted in descending order of importance.
    \item \textbf{Target's Subgraph Nodes' Feature Values and Percentiles:} List of dictionaries with keys as target's subgraph nodes', and values are nodes' feature values and percentile ranks.
    \item \textbf{Target's Subgraph Edge's Influence Importance:} Dictionary with the target node's subgraph edge's influence importance on the target node prediction. The key is a neighbouring node number, and the value represents the importance of the connection of the neighbouring node for the target's final prediction, sorted in descending order.
    \item \textbf{Target's Subgraph Nodes' Labels:} Dictionary with the target's subgraph nodes' labels, where the key is a node index in a subgraph, and the value is the node's label.
    \item \textbf{Model's Prediction:} Dictionary with the target node index and predicted class.
\end{itemize}
- Machine Learning Task Information: \verb|{ml_task_info}|

- Dataset Information: \verb|{dataset_info}|

- Target Node's Data: \verb|{tardet_nodes_data}|

- Target Node's Categorical Data: \verb|{tardet_nodes_data_cat}|

- Target Node's Number of Edges and Labels Ratio: \verb|{nodes_number_of_edges_and_labels_ratio}|

- Target's Subgraph Nodes' Feature Values and Percentiles: \verb|{subg_nodes_features_and_percentiles}|

- Target's Subgraph Edge's Importance Weights: \verb|{subg_edges}|

- Target's Subgraph Nodes' Labels: \verb|{subg_nodes_labels}|

- Model's Prediction: \verb|{gnn_prediction}|
\end{tcolorbox}
  \caption{The prompt schema used to generate GraphXAINs.}
  \label{fig:prompt}
\end{figure}

\begin{figure}[htbp]
  \centering
\begin{tcolorbox}[colback=blue!5, colframe=blue!80, title=GraphXAIN Prompt Rules, sharp corners]
Generate a fluent and cohesive narrative that explains the prediction made by the model. In your answer, please follow these rules:

\vspace{1em} \textbf{Format-related rules:}
\begin{enumerate}
    \item Start the explanation immediately.
    \item Limit the entire answer to \verb|{sentence_limit}| sentences or fewer.
    \item Only mention the top \verb|{num_feat}| most important features in the narrative.
    \item Do not use tables or lists, or simply rattle through the features and/or nodes one by one. The goal is to have a narrative/story.
\end{enumerate}

\textbf{Content related rules:}
\begin{enumerate}
    \item Be clear about what the model actually predicted for the target node with index  \verb|{node_index}|.
    \item Discuss how the features and/or nodes contributed to the final prediction. Make sure to clearly establish this the first time you refer to a feature or node.
    \item Discuss how the subgraph's edge importance contributes to the final prediction. Make sure to clearly establish this the first time you refer to an edge connection.
    \item Consider the feature importance, feature values, averages, and percentiles when referencing their relative importance.
    \item Begin the discussion of features by presenting those with the highest feature importance values first. The reader should be able to tell what the order of importance of the features is based on their feature importance value.
    \item Provide a suggestion or interpretation as to why a feature contributed in a certain direction. Try to introduce external knowledge that you might have.
    \item If there is no simple explanation for the effect of a feature, consider the context of other features and/or nodes in the interpretation.
    \item Do not use the feature importance numeric values in your answer.
    \item You can use the feature values themselves in the explanation, as long as they are not categorical variables.
    \item Do not refer to the average and/or percentile for every single feature; reserve it for features where it truly clarifies the explanation.
    \item When discussing a node's categorical data, make sure to indicate whether the presence (1) or absence (0) of a feature is contextually informative and/or significantly contributes to the explanation. State that it is one of the possible values among that category.
    \item When discussing the connections between the nodes, relate how the influence of a node's relationship might impact the final prediction.
    \item When you refer to node and edges, keep in mind that the target node is a \verb|{node_description}| and edges are \verb|{edges_description}| in this dataset.
    \item Tell a clear and engaging story, including details from both feature values and node connections, to make the explanation more relatable and interesting.
    \item Use clear and simple language that a general audience can understand, avoiding overly technical jargon or explaining any necessary technical terms in plain language.
\end{enumerate}

\end{tcolorbox}
  \caption{The prompt rules used to generate GraphXAINs.}
  \label{fig:prompt_cd}
\end{figure}

\newpage
\section{Examples of XAI Explanations in Natural Language for GNN Models}

\subsection{GraphXAIN}
\label{app:Cedro}

\begin{figure}[h!]
    \begin{minipage}{0.47\textwidth}
        \centering
        \includegraphics[width=\linewidth]{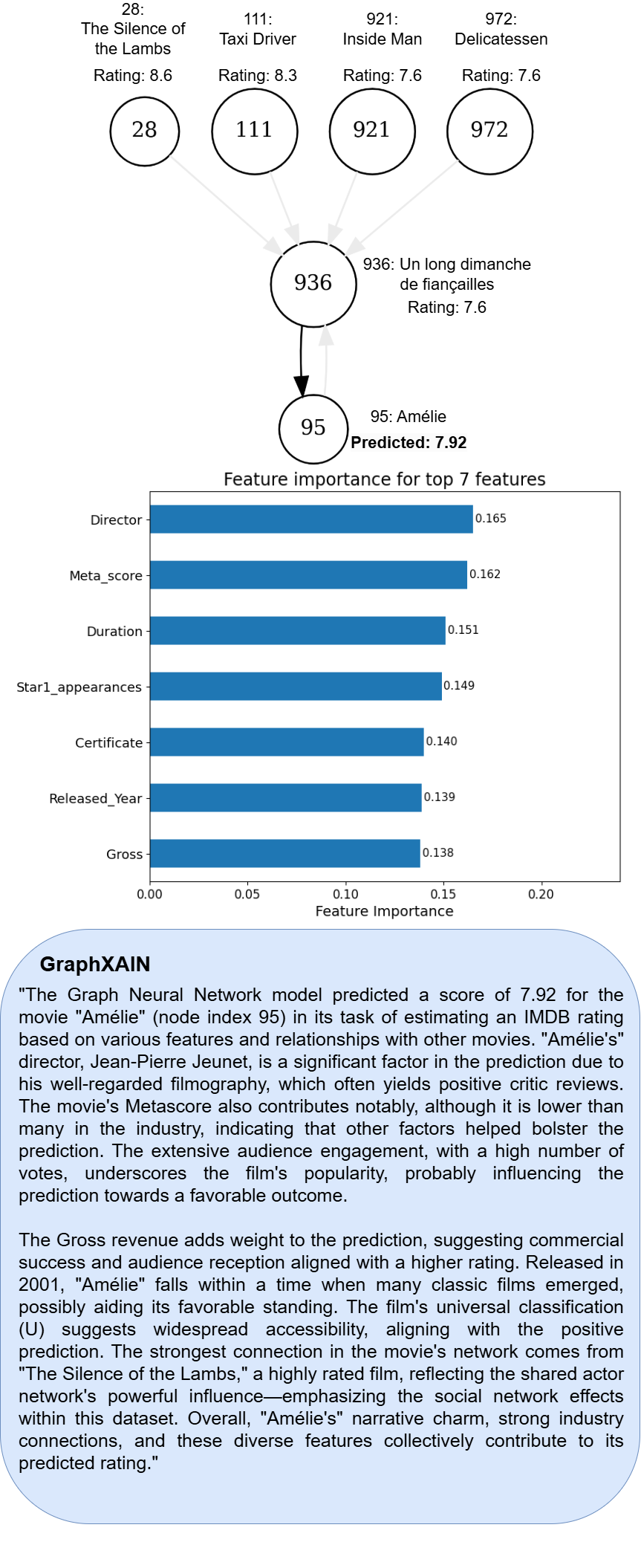}
        \captionof{figure}{LLM-generated GraphXAIN (bottom) complements GNNExplainer’s\cite{gnnexplainergeneratingexplanationsgraph} subgraph and feature importance outputs (top) in explaining the GNN’s prediction for the movie ``Amélie" (node 95) in the IMDB dataset, based on movie features and connections with other movies (nodes) via shared actors.}
        \label{fig:XAIN_IMDB_95}
    \end{minipage}
    \hfill
    \begin{minipage}{0.49\textwidth}
        \centering
        \includegraphics[width=\linewidth]{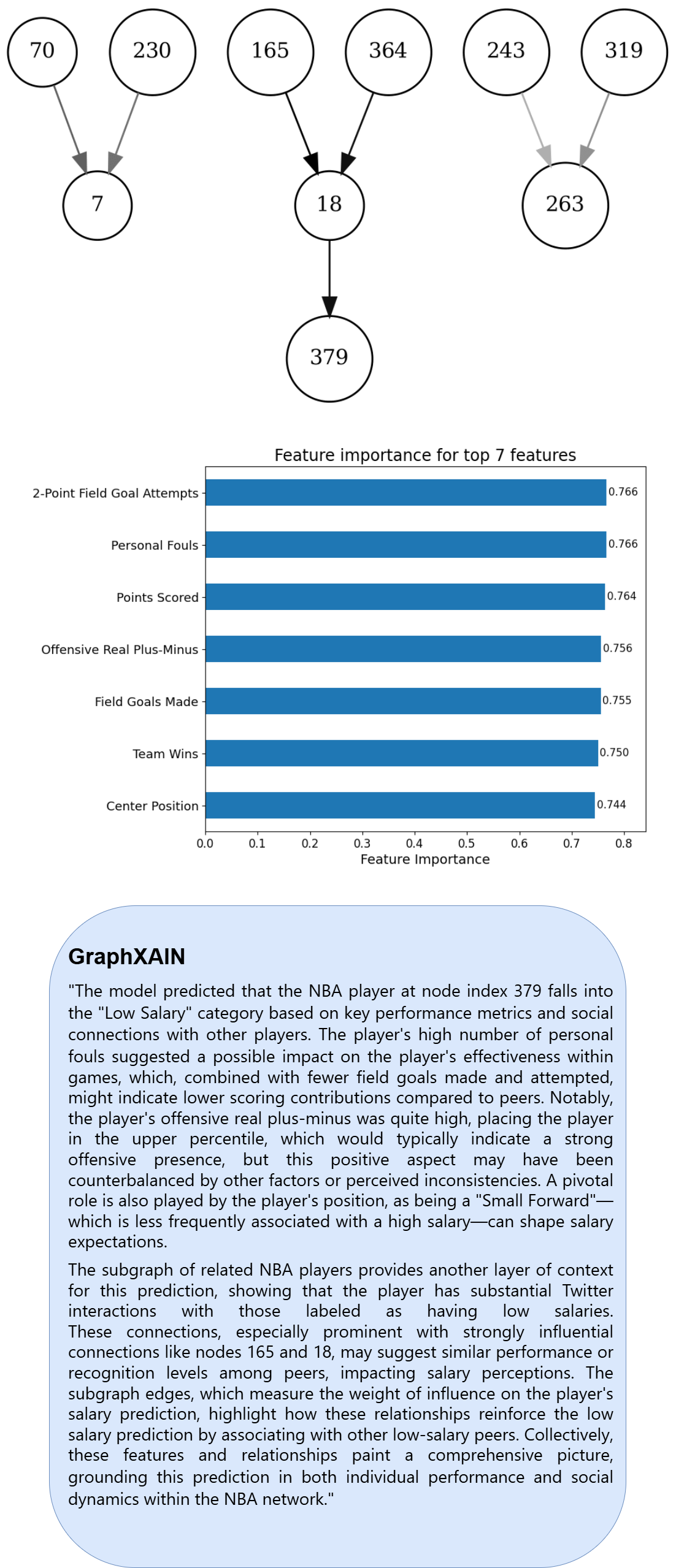}
        \captionof{figure}{LLM-generated GraphXAIN (bottom) complements GNNExplainer's\cite{gnnexplainergeneratingexplanationsgraph} subgraph and feature importance outputs (top) in explaining the GNN's prediction for the player 379 (node 379) in the NBA dataset, based on the player's attributes, field statistics, and social connections with other players (nodes).}
        \label{fig:XAIN_379}
    \end{minipage}
\end{figure}

In this section, we present more examples generated by the GraphXAIN method. Figure \ref{fig:XAIN_IMDB_95} and Figure \ref{fig:XAIN_379} provide further insight into the interpretability of the GNN's model prediction by presenting coherence narratives, which demonstrates the GraphXAIN’s consistency across a range of randomly sampled nodes for both node classification and node regression tasks. Each example was generated following the same methodology, with nodes randomly sampled using the \texttt{random.sample(range(0, len(data.x)), 5)} function and \texttt{set\_seed(42)}. The examples showcase the explanatory depth of GraphXAINs in capturing the reasoning behind the GNN model’s predictions.

In the truncated subgraph visualisations presented in both Figure \ref{fig:XAIN_IMDB_95} and Figure \ref{fig:XAIN_379}, GNNExplainer is instructed to reduce the subgraph to the seven most influential nodes according to edges' weight importance. As a result, this process occasionally results in disconnected nodes within the truncated subgraph, as GNNExplainer prioritises node importance over connectivity in the simplified view. Consequently, these subgraph representations can appear more unintuitive or fragmented. A complementary natural language narrative is therefore argued to be essential to bridge these interpretative gaps, providing end users with a coherent understanding of the GNN model's final prediction, regardless of the connectivity of the subgraph.

\begin{figure}[h!]
    \centering
    \includegraphics[width=0.75\textwidth]{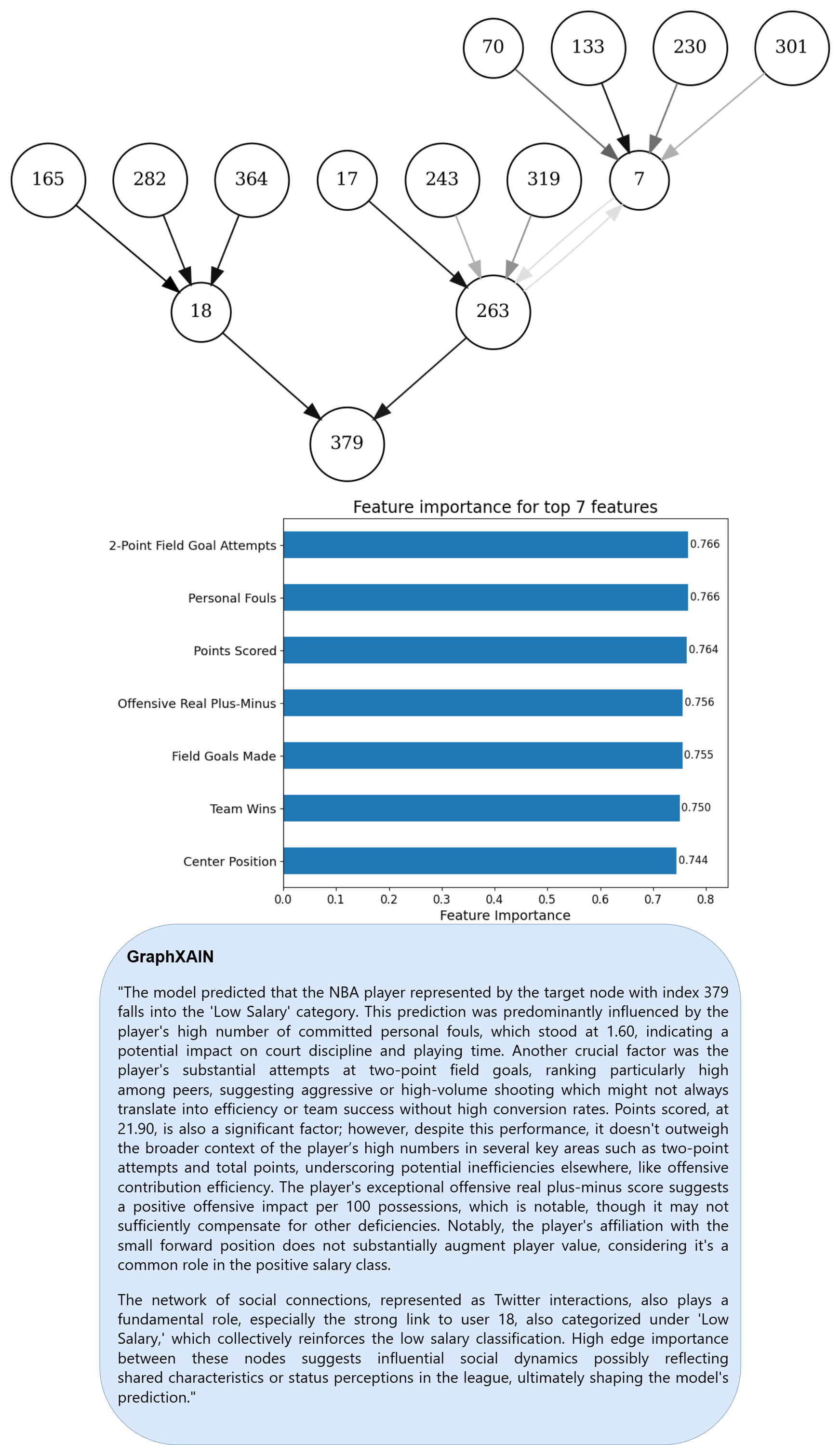}
    \caption{LLM-generated GraphXAIN (bottom) complements the GNNexplainer's outputs (top), displaying a connected subgraph of the 14 most influential nodes for the GNN model’s prediction of node 379. This expanded subgraph presents key connections, including nodes 7 and 263, which significantly impact the final prediction, but also highlights the complexity introduced by a larger subgraph structure.}
    \label{fig:XAIN_379_14}
\end{figure}

For example, to obtain a connected subgraph for node 379, GNNExplainer requires at least 14 nodes in this particular example. In this larger subgraph, previously disconnected nodes 7 and 263 become connected to the target node 379, revealing their substantial influence on the final prediction. However, as the subgraph expands, it becomes increasingly complex and challenging to interpret intuitively. Thus, a complementary natural language narrative is needed to provide a coherent understanding of the GNN model’s prediction. Figure \ref{fig:XAIN_379_14} presents GraphXAIN's output for a connected subgraph containing the 14 most influential nodes for the target node's 379 prediction.

\newpage

\newpage

\newpage
\subsection{Natural Language Counterfactual Explanations for Graphs Using Large Language Models (Giorgi et al.,\ 2024)}
\label{app:Giorgi}

This section presents three examples of textual explanations proposed by Giorgi et al.\ (2024) \cite{giorgi2024naturallanguagecounterfactualexplanations} for counterfactual explanations of node classification using a GNN model trained on the Cora dataset. The Cora dataset is a standard citation network dataset in which nodes represent research papers, and edges denote citation relationships, with the task being to classify each paper into one of seven categories. Since the authors do not provide any output examples in their article, the examples presented below are taken from the article's official GitHub repository. These examples are not selectively chosen but represent the first three explanations generated by the best-performing model, as reported in the paper and available in the article's official GitHub repository\footnote{https://github.com/flaat/llm-graph-cf/blob/main/data/results/cf-gnn\_Qwen2.5-14B-Instruct-GPTQ-Int4\_cora\_Response.json}:

\begin{itemize}
    \item \textit{``The target node (node 15) was originally classified as `Probabilistic\_Methods' in the factual graph. However, in the counterfactual graph, the classification changes to `Reinforcement\_Learning'. Despite the features of node 15 remaining identical, the classification shifts due to changes in the classification of its neighboring nodes. Specifically, nodes 4 and 12, which were previously classified as `Probabilistic\_Methods', are now reclassified as `Reinforcement\_Learning'. This change in the neighborhood context strongly influences the classification of node 15, leading to its reclassification as `Reinforcement\_Learning'. The only neighbor that remains consistent is node 19, but its influence is outweighed by the changes in nodes 4 and 12. Thus, the primary factor driving the change in classification is the altered classification of neighboring nodes, rather than any intrinsic change in the features of node 15 itself."} (Line 4)
    \item \textit{``In the factual scenario, Node 3 is classified as `Theory' and is connected to Nodes 0, 9, 10, and 13. In the counterfactual scenario, Node 3 is reclassified as `Case\_Based' and is only connected to Node 10. The change in classification is influenced by the loss of connections to Nodes 0, 9, and 13, and the retention of a connection to Node 10, which is classified as `Case\_Based'."} (Line 8)
    \item \textit{``In the factual scenario, Node 0 is classified as `Theory' and is connected to both Nodes 1 and 4. Node 1 shares similar features with Node 0 and is also classified as `Theory'. However, Node 4, which is classified as `Case\_Based', introduces a conflicting influence. In the counterfactual scenario, the connection to Node 4 is removed, isolating Node 0 from the `Case\_Based' influence. As a result, Node 0 is reclassified as `Case\_Based', reflecting the reduced influence of `Theory' nodes and the absence of conflicting `Case\_Based' connections. This counterfactual explanation highlights how the removal of a single connection can significantly alter the classification of a node, providing insight into the decision-making process of the graph-based model."} (Line 12)
\end{itemize}

According to Dahlstrom\ (2014) \cite{dahlstrom} and our proposed definitions, these examples constitute an XAI Description rather than an XAI Narrative.

\subsection{Explaining Graph Neural Networks with Large Language Models: A Counterfactual Perspective for Molecular Property Prediction (He et al.,\ 2024)}
\label{app:He}

Presented below are three examples of textual explanations generated by the method proposed by He et al.\ (2024) \cite{he2024explaininggraphneuralnetworks} for counterfactual explanations in the context of molecular property prediction using a Graph Neural Network (GNN) model trained on a chemical molecule dataset. This dataset comprises molecular structures in which nodes represent atoms and edges signify chemical bonds, with the primary objective being the prediction of specific molecular properties. These three examples are the only instances provided by the authors in their article \cite{he2024explaininggraphneuralnetworks}:

\begin{itemize}
    \item \textit{``This molecule contains a cyclohexane ring, a dithiane ring, a ketone group, and a thiocarbonyl group, in which the ketone group may be the most influential for AIDS treatment."} (Page 3, Figure 2)
    \item \textit{``The molecule contains hydroxylamine, cyclohexane, sulfone, and thioether functional groups, in which hydroxylamine may be the most influential for AIDS treatment."} (Page 4, Figure 4)
    \item \textit{``This molecule contains a cyclohexane ring, a dithiane ring, a ketone group, and a hydrazine group, in which the hydrazine group may be the most influential for AIDS treatment."} (Page 6, Figure 5)
\end{itemize}

Again, the explanations presented by He et al.\ (2024) \cite{he2024explaininggraphneuralnetworks} are delivered in a context-free and fact-based manner without illustrating underlying relationships or cause-and-effect relationships, thereby categorising them as XAI Descriptions rather than XAI Narratives.

\subsection{TAGExplainer: Narrating Graph Explanations for Text-Attributed Graph Learning Models (Pan et al.,\ 2024)}
\label{app:Pan}

Presented below are three instances of textual explanations proposed by Pan et al.\ (2024) \cite{TAGex} for Text-Attributed Graph (TAG) learning models using their TAGExplainer method. The authors applied TAGExplainer to the Cora, DBLP, and Book-History datasets to generate natural language explanations that incorporate elements of narrative communication. These three examples are the only instances provided by the authors in their article, each corresponding to the aforementioned datasets, respectively:

\begin{itemize}
    \item \textit{``The classification of the ROOT node into the `Reinforcement Learning' category can be explained as follows:
    \begin{itemize}
    \item ROOT mentions reinforcement learning, learning algorithms, and hierarchical control structures indicating that it focuses on reinforcement learning methodologies and their applications. - Node-1 discusses transfer learning in reinforcement learning, highlighting how learning agents can adapt to multiple tasks.
    \item Node-1.1 - Node-1.8 provide more context about various reinforcement learning architectures and approaches, including modular Q-learning and hierarchical reinforcement learning.
    \item Node-2 focuses on learning in continuous domains with delayed rewards, a pertinent aspect of reinforcement learning, particularly in real-world applications.
    \item Node-2.1 discusses generalization in reinforcement learning using neural networks, which is relevant but less directly tied to the prediction. 
    \item Node-3: This node discusses the need for hierarchies within control systems and the improvements in nested Q-learning, which are highly relevant to the classification label.
    \end{itemize}
    In summary, the classification of the ROOT node into `Reinforcement Learning' is well-supported by the strong presence of key reinforcement learning terms within the node itself and its direct connections to Nodes 1, 2 and 3, which also focus on reinforcement learning methodologies."} (Page 10, Figure 4b)
    \item \textit{``The classification of the ROOT node into the `Cloud computing' category can be explained as follows:
    \begin{itemize}
    \item ROOT: The presence of keywords such as `user', `interfaces', and `service-oriented approach' indicates that the paper is focused on topics central to cloud computing, particularly how users interact with cloud services and the modular nature of these services.
    \item Node-1: While Node 1 discusses `graphical user interfaces' and `domain-specific languages', which are relevant to cloud computing, it is slightly less directly related to the core concepts of cloud computing compared to the ROOT node.
    \item Node-1.1: This node emphasizes `user interface requirements' and `requirements analysis', which are crucial for developing effective cloud applications that meet user needs.
    \item Node-1.2: This node focuses on `stereotypes' and `domain-specific terms', highlighting the importance of understanding user interactions and the context in which cloud services are utilized.
    \end{itemize}
    In summary, the classification of the ROOT node into `Cloud computing' is well-supported by the presence of key terms related to user interaction and service-oriented architectures. The direct connection to Node 1, which discusses user interfaces, further reinforces this classification, while the additional insights from Nodes 1.1 and 1.2 emphasize the importance of user-centric design in cloud computing applications."} (Page 16, Figure 6b)
    \item \textit{``The classification of the ROOT node into the `Europe' category can be explained as follows:
    \begin{itemize}
    \item ROOT: The presence of keywords such as `cambodia', `year', and `translation' indicates a context that may involve historical or cultural discussions relevant to Europe, particularly in terms of colonial history and cultural exchanges. The mention of `english' and `french' highlights the linguistic dimensions that are significant in European contexts.
    \item Node-1: This node discusses Michael D. Coe, an anthropologist specializing in Southeast Asia and the Khmer civilization. While it provides historical context, the focus on Southeast Asia may dilute its direct relevance to Europe. However, the terms `civilizations' and `ancient' could connect to European historical interests.
    \item Node-2: This node is more directly relevant as it discusses the destruction of Cambodia during the Nixon-Kissinger era, a significant historical event that involved European powers' interests in Southeast Asia. The emphasis on `destruction' and `cambodia' alongside key historical figures suggests a critical perspective on the geopolitical dynamics involving European countries.
    \end{itemize}
    In summary, the classification of the ROOT node into `Europe' is supported by the presence of key terms that indicate a historical and cultural context relevant to European interests, particularly through the stronger connection found in Node-2."} (Page 16, Figure 7b)
\end{itemize}
The presented examples align with the proposed XAI Narrative definition. However, the TAGExplainer approach applies to text-attributed graphs only and does not provide an explanatory subgraph or feature importance.

\end{document}